\documentclass[acmsmall, sigconf, screen]{acmart}
\usepackage{transactive-control}

\setcopyright{rightsretained}
\begin{document}
\title{Adapting Surprise Minimizing Reinforcement Learning Techniques for Transactive Control}
\author{William Arnold}
\affiliation{
    \institution{Electrical Engineering and Computer Sciences, University of California, Berkeley}
    \country{}
}
\author{Tarang Srivastava}
\affiliation{
    \institution{Electrical Engineering and Computer Sciences, University of California, Berkeley}
    \country{}
}
\author{Lucas Spangher}
\affiliation{
    \institution{Electrical Engineering and Computer Sciences, University of California, Berkeley}
    \country{}
}
\author{Utkarsha Agwan}
\affiliation{
    \institution{Electrical Engineering and Computer Sciences, University of California, Berkeley}
    \country{}
}
\author{Costas Spanos}
\affiliation{
    \institution{Electrical Engineering and Computer Sciences, University of California, Berkeley}
    \country{}
}

\begin{abstract}
Optimizing prices for energy demand response requires a flexible controller with ability to navigate complex environments. 
We propose a reinforcement learning controller with surprise minimizing modifications in its architecture. 
We suggest that surprise minimization can be used to improve learning speed, taking advantage of predictability in peoples' energy usage. Our architecture performs well in a simulation of energy demand response. We propose this modification to improve functionality and savin a large-scale experiment.   
\end{abstract}


\copyrightyear{2021}
\acmYear{2021}
\acmConference[e-Energy '21]{The Twelfth ACM International Conference on Future Energy Systems}{June 28-July 2, 2021}{Virtual Event, Italy}
\acmBooktitle{The Twelfth ACM International Conference on Future Energy Systems (e-Energy '21), June 28-July 2, 2021, Virtual Event, Italy}\acmDOI{10.1145/3447555.3466590}
\acmISBN{978-1-4503-8333-2/21/06}


\maketitle

\section{Introduction} \label{sec:intro}
The electricity grid may be seen as a beautifully decentralized organism: one in which individual energy demands are met by individual generators largely without direct coordination or knowledge of recipient. 
Market signals simply translate into generators commanding their resources to increase or decrease. 
However, as volatile resources like wind and solar replace on-demand resources like fossil fuels, a potentially worrying question arises: what happens to demand when the generation becomes decoupled from commands; 
i.e., when energy is demanded but the sun isn't shining? 
Grids that do not adequately prepare for this question will face daunting consequences, 
ranging from curtailment of resources \citep{spangher2020prospective} to voltage instability and physical damage.  

A common solution is demand response: a strategy in which customers are incentivized to shift loads to periods of the day where energy is plentiful. 
Given the lack of material infrastructure required, it has several positives above physical energy storage systems. 
One primary application of demand response is in buildings, and central-level appliance coordination has been thoroughly studied in residential and industrial settings
\citep{asadinejad2018evaluation,ma2015cooperative,li2018integrating,yoon2014dynamic,johnson2015dynamic}.
However, while physical infrastructure of office buildings has been studied for demand response (\citep{8248801}), 
there has been no large scale experiment aimed to elicit a behavioral demand response. 

The lack of experiment is understandable given most offices do not have mechanisms to pass energy prices to office workers \citep{das2019novel}. 
If they did, however, not only could a large fleet of decentralized batteries 
-- laptops, cell phone chargers, etc. -- 
be coordinated to function as a large deferable resource, 
but building managers could save money\citep{das2020occupants}.

The SinBerBEST collaboration has developed a Social Game that facilitates workers to engage in competition around energy \citep{konstantakopoulos2019deep}, \citep{konstantakopoulos2019design}. 
Through this framework, a first-of-its-kind experiment has been proposed to implement behavioral demand response within an office building \citep{spangher2020prospective}. 
Prior work has proposed to describe an hourly price-setting controller that learns how to optimize its prices \citep{spangher2020augmenting}. 

However, given the costliness of iterations in this experiment, 
further work in simulation is needed to refine a controller that can adapt faster to the behavior of office workers.  

We endeavor to report one such refinement. 
In Section \ref{sec:background} we will contextualize the architecture of our reinforcement learning (RL) controller within the general domain of RL and improve it with a surprise minimizing algorithm.
In Section \ref{sec:methods} we will describe the simulation setup and the specific modification we propose. 
In Section \ref{sec:results} we will give results and in Section \ref{sec:discussion} we will discuss implications of the controller and the future work this entails. 

\section{Background} \label{sec:background}

\subsection{Reinforcement Learning}
Reinforcement learning (RL) is a type of agent-based machine learning 
where control of a complex system requires actions that maximize the agent's outcome \citep{sutton2018reinforcement}, 
i.e. they seek to optimize the expected sum of rewards for actions ($a_t$) and states ($s_t$) with a policy $p_\pi$ and reward given by $ r $ 
in a policy parametirized by 
$\theta$; i.e.,  $J(\theta) = \E\sum_{s_t,a_t \sim p_{\pi}}[r(s_t, a_t)]$. 

Policy gradient methods are a class of RL algorithms used to train policy networks that suggest actions.
We propose the use of a variant of these methods, Proximal Policy Optimization (PPO) \citep{schulmanPPO}, which works by computing an estimator of the policy gradient 
and plugging it into a stochastic gradient ascent algorithm. 

RL has been used for a number of demand response situations, but most of the work consists of agents that directly schedule resources
\citep{6963416,7018632,6848212,6915886,RAJU2015231,FUSELLI2013148}.
RL architectures can vary significantly, 
for example Mbuwir et. al. manages a battery directly using batch Q-learning \citep{en10111846}. 
In another example, Kofinas et. al. deploys a fuzzy Q-learning multi-agent that learns to coordinate appliances to increase reliability \citep{KOFINAS201853}.

\subsection{Surprise Minimizing Reinforcement Learning}

Some stability of control is desirable, for both human subjects and the agent. 
Incentivizing the agent to minimize the surprise it experiences is equivalent to incentivizing it to minimize peoples' change in energy usage across days. This corresponds to adjusting people's habits in a stable system rather than forcing them to confront and attempt to understand an unstable one.
Additionally, people behave predictably on aggregate, and thus choosing to minimize surprise may in fact make it easier for the agent to learn.
In general, we define surprise minimization to be the reduction of novel events experienced by the RL agent. In this context, this equates to the RL controller experiencing stable energy demands by the office workers. 

Surprise Minimizing Reinforcement Learning (SMiRL) is an algorithm that aims to reduce 
the entropy of visited states. SMiRL is useful when the environment 
provides sufficient unexpected and novel events for learning 
where the challenge for the agent is to maintain a 
steady equilibrium state. \citep{smirl}

SMiRL maintains a distribution $ p_{\theta}(s) $ about which states are likely under its current 
policy. The agent then modifies its policy $ \pi $ so that it encounters states $ s $ 
with high $ p_{\theta}(s) $, as well as to seek out states that will change the model $ p_{\theta}(s) $ so 
that future states are more likely. 

We make use of SMiRL as an auxiliary reward in addition to our 
usual reward to calculate a combined reward
$ r_{\text{combined}} = r_{\text{energy}} + \alpha r_{\text{SMiRL}} $. 
With a SMiRL weight $ \alpha $ as a measure of how much the SMiRL reward influences the total reward. 
We will describe the explicit formulation of the SMiRL reward in Section \ref{sec:methods}. 

\section{Methods}\label{sec:methods}

We adapt SMiRL to the problem of optimizing a price-serving agent for energy demand response. 
We will briefly explain the baseline Proximal Policy Optimization (PPO).
We will then describe the simulation environment we test this in. 

\subsection{Environment}

\begin{figure}
\centerline{
\includegraphics[width=0.9\linewidth]{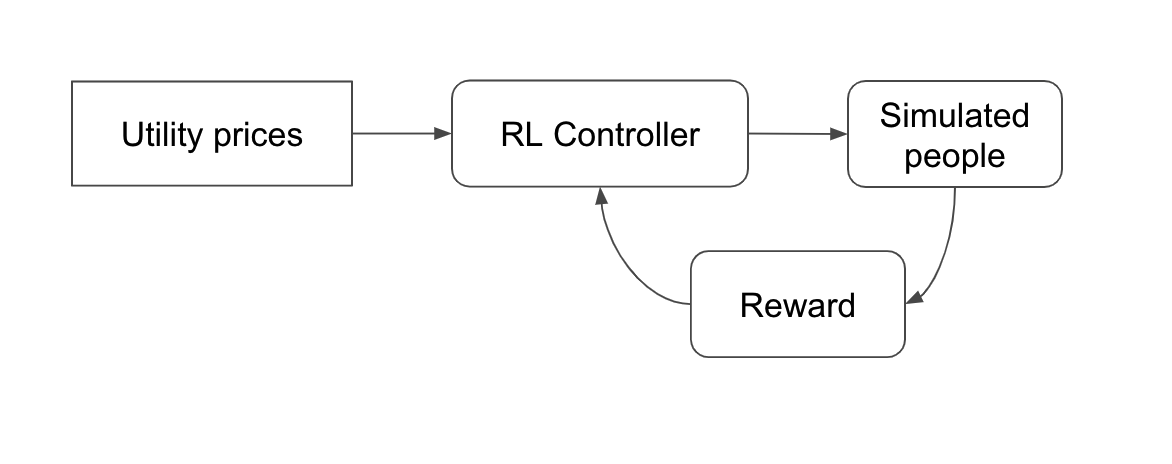}
}
\caption{Reinforcement Learning Control Flow} \label{fig:RLcontroller}
\end{figure}

We summarize an OpenAI gym environment modeled after an environment to simulate demand response in office buildings, visualized in Figure \ref{fig:RLcontroller} \cite{openai,spangherofficelearn}.  
Each step in the environment is a day, where the RL controller (or agent) is given prices by the utility. 
The controller then proposes prices to office workers, who modify their behavior in order to achieve the lowest cost of energy possible, which is then rewarded by the controller. 
The controller then aims to assign prices to office workers which minimize the total cost of energy paid to the utility. 

Can an RL agent learn to provide optimal price signals by implicitly predicting causal factors? 
The space of reasonable energy prices to assign is simple enough to efficiently cover with an entropy maximizing agent, so we employ a Proximal Policy Optimization (PPO) architecture. 
We use Ray's RLLib PPO algorithm.
The reward for the price-setting agent is 
$ r_{\text{energy}} = \log(d^tg) $
where $d$ is the demand of the person it studies, and $g$ is the grid pricing. 
We use parameters of learning rate $ 
\alpha = 0.003$, batch size of $ 256 $, training stochastic gradient descent 
minibatch of $ 32 $, clip parameter $ 0.3 $ and all other parameters were RLLib
PPO defaults. 
For other implementation choices, please see our Github\footnote{Our Github may be found at the following link: \url{https://github.com/Aphoh/temp_tc}.}

\subsection{SMiRL Reward Formulation}
A SMiRL agent recieves an auxiliary reward for 
experiencing familiar states based on an updating distribution of states it has experienced. 
This is exactly equivalent to learning a policy with the lowest entropy. 
Assuming we have a fully-observed controlled Markov process (CMP) with state $ s_t $ and action $ a_t $
at time $ t $, and $ p(s_0) $ as the initial state distribution, and transition probabilities 
$ T(s_{t+1} | s_t, a_t) $, the agent learns a policy $ \pi_\phi (a | a) $ parameterized by 
$ \phi $. 
As described earlier, we keep track of an estimated state marginal $ p_{\theta_{t-1}} (s_t) $ 
for the actual state marginal $ d^{\pi_\phi} (s_t) $. 
As usual we denote entropy of a state $ s_t $ by $ \entropy{s_t} $. 
The entropy can then be calculated by the marginal as 
\begin{align}
    \sum_{t=0}^{T} \mathcal{H}\left(\mathbf{s}_{t}\right)
    &= -\sum_{t=0}^{T} \mathbb{E}_{\mathbf{s}_{t} \sim d^{\pi} \phi\left(\mathbf{s}_{t}\right)}\left[\log d^{\pi_{\phi}}\left(\mathbf{s}_{t}\right)\right] \\
    &\leq-\sum_{t=0}^{T} \mathbb{E}_{\mathbf{s}_{t} \sim d^{\pi} \phi\left(\mathbf{s}_{t}\right)}\left[\log p_{\theta_{t-1}}\left(\mathbf{s}_{t}\right)\right] 
\end{align}
We bound (1) by the entropy of an estimated marginal $ p_{\theta_{t-1}} $ in (2).
Minimizing the right side bound is then equivalent to maximizing an RL objective with reward 
$ r_{\text{SMiRL}}(s_t) = \log p_{\theta_{t-1}} (s_t)  $
We note that the optimal policy must also consider future changes to $ p_{\theta_{t-1}} (s_t) $
since the distribution of visited states changes at each step. 
To account for this we use an augmented MDP that captures this notion. \citep{smirl}
We note that in our implementation of SMiRL, $ p_{\theta_t} (s) $ is normally distributed. 
To construct the augmented MDP we include sufficient statistics for $ p_{\theta_t} (s)$ 
in the state space such as the paramaeters of our normal distribution and the number of states 
seen so far. 

\subsection{SMiRL Implementation}
SMiRL is simply implemented in our existing OpenAI socialgame environment. 
We introduce SMiRL into our existing socialgame environment by initializing a 
buffer that tracks the agent's observed states and computes an estimated state marginal $ p_{\theta_t} $. 
As noted earlier, in the augmented MDP the state space also contains the number of observed states and this information is stored in the buffer as well. 
At each step in our simulation we add newly observed states to our buffer, and update $p_{\theta_t}$. The agent then adjusts the its policy based on the combined reward. 

We notice that since $ p_{\theta}(s) $ is modeled as an independent Gaussian for each dimension (hour) in the observation (consumption for a day), 
then the SMiRL reward is expressed as 
\begin{align}
    r_{\text{SMiRL}}(s_t) = - \sum\limits_i \left( \log \sigma_i + \frac{(s_i - \mu_i)^2}{2\sigma_i^2} \right) 
\end{align}
where $ \mu_i $ and $ \sigma_i$ are the sample mean and standard deviation from our state marginal 
and $ s_i $ is the $ i^{th}$ feature ($ i^{th} $ hour of day) of $ s $ \citep{smirl}.
With this formulation we can efficiently calculate the SMiRL reward from our buffer. 

\subsection{SMiRL as an Auxilary Reward}
We use SMiRL as an auxilary reward to provide faster learning and more stable outputs. 
We achieve this by calculating the SMiRL reward $ r_{\text{SMiRL}} $ as described in the previous section, and applying a SMiRL weight $ \alpha $ to it and then using the sum with our usual energy 
reward $ r_{\text{energy}} = \log(d^tg) $. This gives us a combined reward of 
\begin{equation}
    r_{\text{combined}} = r_{\text{energy}} + \alpha r_{\text{SMiRL}}
\end{equation}
and it is this reward that we use to train the RL agent. 
In our simulations, we found the optimal SMiRL weight $ \alpha $  to be around $ \alpha = 0.12$ after 
hyperparameter tuning. 
We will discuss the exact results of various SMiRL weights in Section \ref{sec:results}. 

\section{Results}\label{sec:results}
We now discuss our two aims: more effective learning and a more stable environment.

\subsection{Sample Entropy}
\subsubsection{Observing Lower Sample Entropy}
As the primary objective of a surprise minimization technique is to, in fact,  minimize surprise, it is natural to determine whether SMiRL does so.

\begin{figure}
    \centering
    \includegraphics[width=\linewidth]{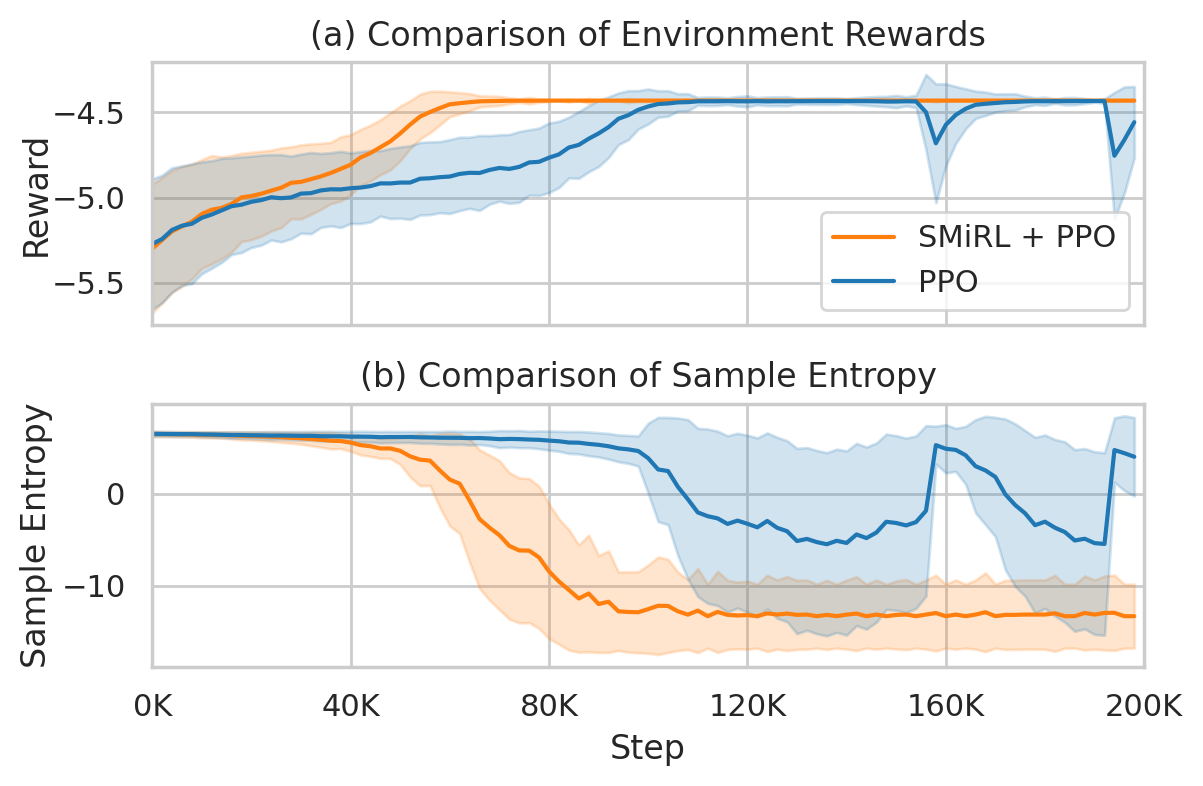}
    \caption{A comparison between the PPO + SMiRL agent and the baseline PPO agents' (a) rewards and (b) sample entropies over training steps. Shaded regions are one standard deviation of observations binned to every 100 steps.}
    \label{fig:sample_entropy}
\end{figure}

We first quantify the degree to which the environment is more or less entropic when under the influence of an agent employing SMiRL. Since we assume each variable in our sample space is normally distributed in a given timeframe, we can compute its entropy as $ H = \frac{1}{2} \ln (2 \pi e \sigma^2)$. For each step, we compute the sample variance of the last 100 steps and use this to compute a sample entropy, shown in \textbf{Figure \ref{fig:sample_entropy}}. 

While both agents do reduce the sample entropy over time, the SMiRL + PPO agent does so earlier than the baseline PPO agent. 
The SMiRL + PPO implementation exhibits lower sample entropy for all iterations compared to baseline PPO, and hence energy usage over the course of the simulation is more consistent when using SMiRL.

Additionally, the SMiRL + PPO agent converges towards a policy which generates a more stable final environment than the baseline PPO agent. 
The SMiRL + PPO agent's behaviour confirms our hypothesis that the stability of our environment is superior to our baseline PPO agent by incentivizing the RL agent to revisit familiar states. 
We also note that the both agents continue to explore changes to their policies after convergence, however the SMiRL + PPO maintains a lower sample entropy in late timesteps. Hence, the SMiRL + PPO agent explores in ways that maintain a more stable environment, while the PPO agent's exploration results in more unstable energy usage.

In this sense, the addition of the SMiRL reward can allow an agent to strike a balance between exploration and stability. 

\subsection{Improved Learning with SMiRL (Compared to Baseline PPO)}
While environment stability is important, it is essenital that the agent still encorages efficient energy usage. We also wish to understand whether the SMiRL reward models our assumption of predictability in our system by improved learning speed.

\subsubsection{Faster Learning and Consistent Outcomes}\label{sec:RL_learning}
We observe that our SMiRL + PPO implementation induces significantly faster learning and convergence to an equally optimal policy, with the same reward. 
As shown by Figure \ref{fig:sample_entropy}(a), both agents maintain similar rewards up until step \textasciitilde 30k, after which the SMiRL + PPO agent begins to achieve, on average, a higher reward.
The SMiRL + PPO implementation converges in roughly half the time as the baseline PPO agent (step \textasciitilde 50k v.s. step \textasciitilde 110k). The reward in the environment around step \textasciitilde 60k whereas the PPO baseline does not 
converge to the maximum reward in the environment until step \textasciitilde 110k. 
Our results support our hypothesis that an auxilary SMiRL reward improves learning speed. 

Note that we compare the energy rewards ($r_\text{energy}$) here directly and not the combined reward which includes the SMiRL reward and weight. This demonstrates the influence of the SMiRL reward on the energy reward, which describes the overall effectiveness of our agent. Hence the inclusion of the SMiRL reward results in improved learning speed of our agent in the task it is given. 
\begin{figure}
    \centering
    \includegraphics[width=\linewidth]{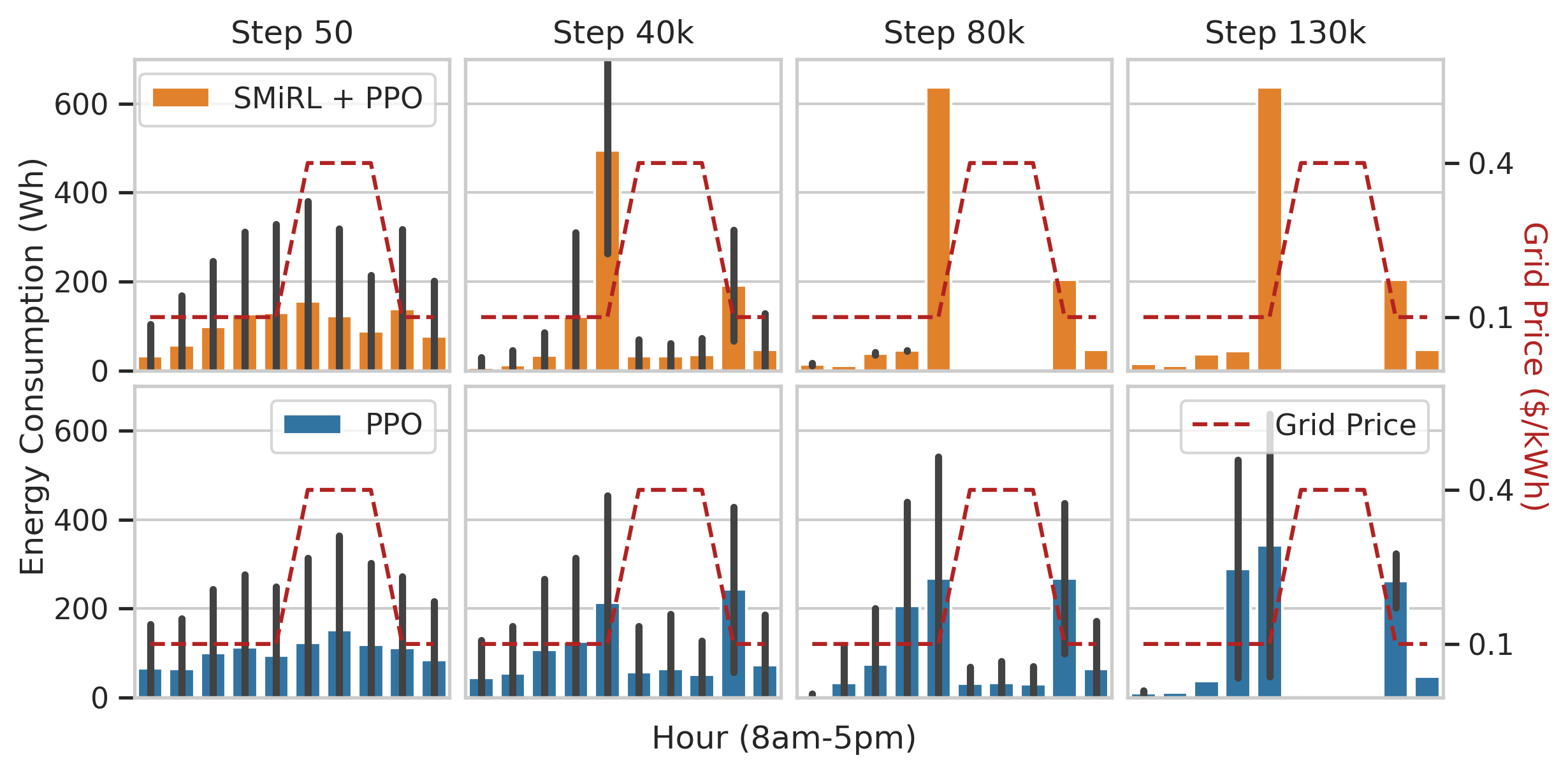}
    \caption{Energy consumption with the PPO and SMiRL + PPO agent at steps 10k, 40k, 80k, and 130k compared to the grid price. Please see section \ref{sec:discussion} for a brief note on the seemingly adverse behavior in the SMiRL+PPO.}
    \label{fig:observations}
\end{figure}

Faster convergence is demonstrated by Figure \ref{fig:observations}, where energy consumption is shown at different steps in the environment.
The grid price signal the agent recieves is shown. We observe that, for both the PPO and SMiRL + PPO agent, the price signal  effectively shifts people's energy consumption away from times when the grid price is higher.
By step 40k, however, the SMiRL + PPO agent has already begun to greatly reduce consumption during peak, and people have shifted it towards just before that peak.
By step 80k, the SMiRL + PPO agent has completely diminished any energy usage during the peak pricing, while the PPO agent only does so by step \textasciitilde 120k (when its reward converges).

\subsubsection{Optimal SMiRL Weights}
We note that while an appropriate SMiRL weight provides significant improvement in learning speed
and sample entropy, an inadequate SMiRL weight can lead to poor learning that converges to a suboptimal result, and may not outperform a baseline PPO. 

Specifically, we found SMiRL weights of $ \alpha = 0.25$ and higher performed worse than baseline PPO. 
In fact, we find that they converge to a suboptimal reward in the environment, hinting that too much surprise minimization might hinder exploration. 

For much lower SMiRL weights such as $ \alpha = 0.01 $, we do not see any 
significant benefits when compared to baseline PPO; there isn't enough 
weight on the SMiRL reward to have a meaningful impact. 

\subsubsection{Sample Entropy and Environment Reward Curves}
When comparing the sample entropy and reward curves in \textbf{Figure \ref{fig:sample_entropy}} 
we see that beginning at steps \textasciitilde 50k, the SMiRL + PPO agent's observed 
sample entropy drops significantly below that of the baseline PPO agent. 
This correlates closely at \textasciitilde 50k steps where the PPO + SMiRL implementation begins to experience significantly greater rewards than baseline PPO. 
Lastly, we observe that at \textasciitilde 110k steps, both agents have observed a drop in sample entropy, and their environment rewards converge.
This correlation between entropy and reward may support our hypothesis that aiming for a stable environment via surprise minimization can help the agent learn faster.


\section{Discussion} \label{sec:discussion}

A careful observer may note in Figure \ref{fig:observations} that although the energy consumptions are certainly minimizing the price that a building manager might pay (i.e. the controller's reward), large peaks would be unfavorable for the grid if incentivized large scale. We argue that this is not a defect perse of the SMiRL modification; only the exploitation of an externality our simulation does not account for. First, the SMiRL agent is achieving a goal: environment stability, evidenced in the tightening of confidence bounds around energy consumption outcomes. Outcome behavior confidence is accomplished by incentivizing people's consumption tightly up against the shoulder of TOU pricing because the mechanics of the simulated office worker cause it to shift energy usage when offered high prices. Therefore the controller learns to rely on shifting. While the SMiRL + PPO and PPO agents are equal in the energy reward the controller receives, the inclusion of the SMiRL reward creates more consistent energy usage from day to day. The behavior is optimal given the externalities we have encoded (or not encoded) into the reward, so should not be seen as a deficit of the controller. It would be straightforward, and the subject of future work, to modify the reward in order to direct the controller to spread energy evenly amonst non-peak hours. 

After consider this, SMiRL has shown promising results in increasing environment stability, 
and observed improvements in learning speeds 
support our hypothesis that SMiRL can take advantage of inherent predictability in people's 
energy consumption. 

From the energy consumptions in Figure \ref{fig:observations} we note that SMiRL induces faster convergence
to cheaper energy consumptions, which would translate into direct energy cost.


The results of our simulation and related work show that in appropriate environments, the 
auxiliary SMiRL reward can improve learning while also letting the RL agent explore. We argue that a human-in-the-loop environment is necessarily complex enough that encouraging surprise minimization encourages the agent to more quickly grasp the latent similarities that govern the environment. We look towards our experiment to continue to demonstrate this. 

\subsection{Future Research}
SMiRL can handle variations in a $k$-discrete state-space by creating $k$ separate buffers, one for each variation of the state-space, that shrink the action sampling towards each $k$ category. We propose to adapt this to a continuous state-space by clustering main categories of observation. One of the main inputs to the state space is price signals, so we propose to cluster based on a semantic feature search proposed in \citep{afzalan2019semantic} for building load profiles. 


\subsection{Acknowledgements}
We gratefully acknowledge the input of Manan Khattar, Austin Jang, and Dustin Luong as working group collaborators, and Glen Berseth for his initial work on SMiRL.

\bibliographystyle{ACM-Reference-Format}
\bibliography{citations}

\end{document}